\documentclass[final]{cvpr}

\usepackage{times}
\usepackage{epsfig}
\usepackage{graphicx}
\usepackage{amsmath}
\usepackage{amssymb}

\usepackage{algorithm}
\usepackage{algorithmic}

\usepackage{booktabs}
\usepackage{diagbox}
\usepackage{stackengine}
\usepackage{color}

\newcommand\ChangeRT[1]{\noalign{\hrule height #1}}
\newcommand\xrowht[2][0]{\addstackgap[.5\dimexpr#2\relax]{\vphantom{#1}}}

\usepackage[pagebackref=true,breaklinks=true,colorlinks,bookmarks=false]{hyperref}

\def\cvprPaperID{2813} 
\def\confYear{CVPR 2021}

\begin{document}

\title{MetaCorrection: Domain-aware Meta Loss Correction for Unsupervised Domain Adaptation in Semantic Segmentation}

\author{Xiaoqing Guo$^{1*}$~~~~~Chen Yang$^{1}$\thanks{Xiaoqing Guo and Chen Yang contributed equally.}~~~~~Baopu Li$^{2}$~~~~~Yixuan Yuan$^{1}$\thanks{This work was supported by Shenzhen-Hong Kong Innovation Circle Category D Project SGDX2019081623300177 (CityU 9240008).}
\\
$^{1}$City University of Hong Kong~~~~~~$^{2}$Baidu USA\\
{\tt\small \{xqguo.ee, cyang.ee\}@my.cityu.edu.hk}~~~~~{\tt\small baopuli@baidu.com}~~~~~{\tt\small yxyuan.ee@cityu.edu.hk}
}

\maketitle

\begin{abstract}
Unsupervised domain adaptation (UDA) aims to transfer the knowledge from the labeled source domain to the unlabeled target domain. Existing self-training based UDA approaches assign pseudo labels for target data and treat them as ground truth labels to fully leverage unlabeled target data for model adaptation. However, the generated pseudo labels from the model optimized on the source domain inevitably contain noise due to the domain gap. To tackle this issue, we advance a MetaCorrection framework, where a Domain-aware Meta-learning strategy is devised to benefit Loss Correction (DMLC) for UDA semantic segmentation. In particular, we model the noise distribution of pseudo labels in target domain by introducing a noise transition matrix (NTM) and construct meta data set with domain-invariant source data to guide the estimation of NTM. Through the risk minimization on the meta data set, the optimized NTM thus can correct the noisy issues in pseudo labels and enhance the generalization ability of the model on the target data. Considering the capacity gap between shallow and deep features, we further employ the proposed DMLC strategy to provide matched and compatible supervision signals for different level features, thereby ensuring deep adaptation. Extensive experimental results highlight the effectiveness of our method\footnote{\url{https://github.com/cyang-cityu/MetaCorrection}} against existing state-of-the-art methods on three benchmarks.
\end{abstract}

\vspace{-0.5cm}
\section{Introduction}
\vspace{-0.2cm}
Unsupervised domain adaptation (UDA) aims to adapt a model for the unlabeled target domain through transferring the knowledge from a labeled source domain with the same label space. UDA for semantic segmentation is a crucial practical problem since it may be beneficial for various real-world applications, such as simulation for robots \cite{james2019sim} and autonomous driving \cite{wang2020differential}. The main challenge of UDA semantic segmentation lies in the divergence of data distribution between two domains \cite{chen2020adversarial, zheng2020rectifying}. Such domain gap often results in significant performance degradation if the model learned on the labeled source data is directly applied to the target samples \cite{zhang2019category, zou2018unsupervised}. 

There exist two major lines of approaches to tackle the domain gap problem. On one hand, adversarial learning based UDA methods as a dominant stream have been devised to bridge the domain gap by aligning the distributions of two domains in the appearance \cite{chen2019learning, chen2019crdoco, choi2019self, li2019bidirectional}, feature \cite{chen2019synergistic, du2019ssf, tran2019gotta, xu2020adversarial} or output spaces \cite{kim2020learning, luo2019taking, tsai2018learning, tsai2019domain}. Despite the significant progress of domain alignment, these works ignored the domain-specific knowledge and could not guarantee the sufficient discriminative capability of the classifier for the specific task. On the other hand, self-training based methods \cite{chen2019domain, chen2020adversarial, iqbal2020mlsl, lian2019constructing, liang2019exploring, mei2020instance, pan2020unsupervised, sakaridis2019guided, wang2020unsupervised, zhang2019category, zheng2020rectifying, zhu2020improving, zou2019confidence, zou2018unsupervised} have emerged as  promising alternatives towards UDA, which enhanced the discrimination property of target features and implicitly encouraged cross-domain alignment by simultaneously training with pseudo-labeled target data and labeled source data. Specifically, self-training methods assign pixel-wise pseudo labels according to confidence score \cite{iqbal2020mlsl, lian2019constructing, pan2020unsupervised, zou2018unsupervised} or uncertainty \cite{liang2019exploring, zheng2020rectifying}, providing extra supervision for target data to optimize the model. 

\begin{figure}[t]
\centering
\includegraphics[width=83mm]{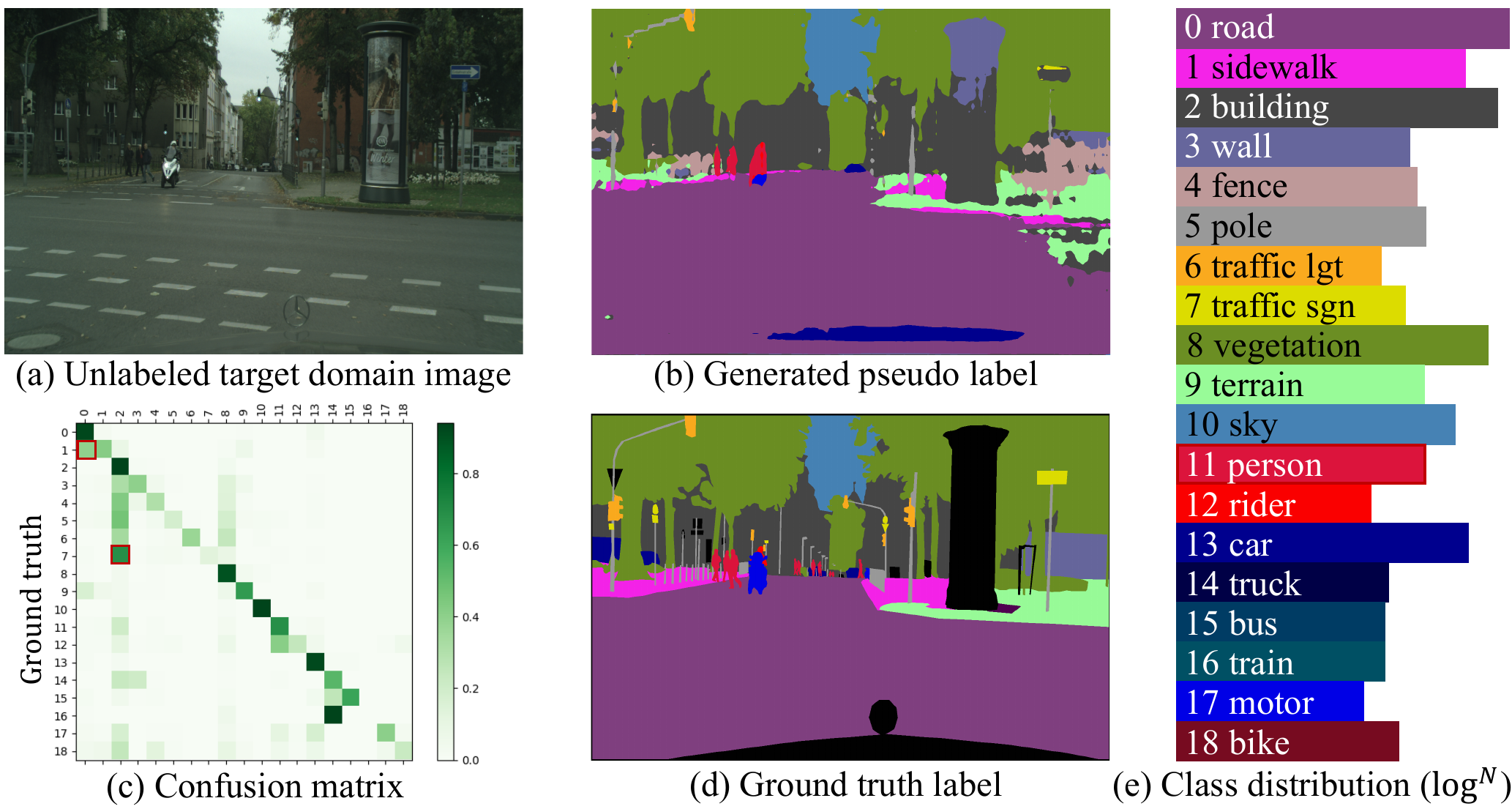}
\caption{Sample of the noisy pseudo labels on Cityscapes \cite{cordts2016cityscapes}. The generated pseudo labels suffer from the data distribution biases in comparison to the ground truth.}
\vspace{-0.4cm}
\label{fig:Motivation}
\end{figure}

However, one issue with self-training based UDA methods is that the generated pseudo labels usually suffer from the noise problem, as illustrated in Figure \ref{fig:Motivation} (b). The presence of noisy pseudo labels may severely hamper the generalization ability of the adapted models, because deep neural networks (DNN) may overfit due to these noisy labeled data. Although some existing works \cite{guo2020semi, lian2019constructing, zou2018unsupervised} manually define a threshold to eliminate the low-confidence pseudo-labeled samples, it is still challenging in several aspects. First, the threshold value is hard to predefine manually. It may depend on many factors such as the stage of training procedure, the degree of discrepancy between two domains, the number of pixels in each class, the location of the pixel, etc. Secondly, those selected training samples may be misclassified with high confidence, leading to accumulated errors. In fact, the noisy pseudo labels tend to appear in under-represented minor classes or ambiguous classes. For instance, the minor category `traffic sign' is overwhelmed by major category `building', and `road' is usually connected to `sidewalk', yielding noisy labels, as in Figure \ref{fig:Motivation} (c). In this scenario, noisy pseudo labels can be theoretically converted from the ground truth labels via a NTM \cite{shu2020meta, zhang2020distilling}, which encodes the inter-class misclassification relationship. 

To heuristically discover intrinsic inter-class noise transition probabilities underlying target data, we model the noise distribution of pseudo labels by a NTM and devise a domain-aware meta-learning strategy to estimate the NTM in a learning-to-learn fashion. The key idea of domain-aware meta-learning is to obtain the meta-knowledge of underlying label distribution of clean data in the target domain, and we introduce a domain predictor to adaptively select domain-invariant source data with ground truth labels as meta data set, so as to guide the derivation of NTM. The domain-aware meta-learning strategy enables the gradient of empirical risk measured on meta data to update the NTM, thereby boosting the generalization capacity. Then the approximated noise distribution can be utilized to explicitly correct the supervision signal for target data, aiming at solving the noisy pseudo label problem in a self-training based UDA method. An alternating optimization approach is further adopted to mutually improve the estimation of NTM and the UDA segmentation model. For simplicity, we refer to the whole Domain-aware Meta-learning strategy for Loss Correction in the above process as DMLC. Moreover, we devise a MetaCorrection framework, which incorporates DMLC to provide matched supervision signals for outputs of different levels, thereby enhancing the deep adaptation of model. In particular, we introduce the learnable NTMs for different layers, and adopt the proposed domain-aware meta-learning to estimate the corresponding noise distributions and benefit the loss correction.  

We summarize our contributions in four aspects. 
\begin{itemize}
\vspace{-0.3cm}
  \item We present a MetaCorrection framework, which incorporates the proposed DMLC strategy for UDA semantic segmentation. To our best knowledge, it represents the first effort to formally model the noise distribution of pseudo labels in target domain by a learnable NTM and further solve it in a meta-learning strategy.
\vspace{-0.3cm}
  \item In the DMLC strategy, we formulate the misclassification probability of inter-classes to model noise distribution in target domain and devise a domain-aware meta-learning algorithm to estimate NTM for loss correction in a data driven manner. 
\vspace{-0.3cm}
  \item Our MetaCorrection framework aims to provide matched and compatible supervision signals for different layers with the proposed DMLC strategy, boosting the adaptation performance of model. 
\vspace{-0.3cm}
  \item We conduct extensive comparison experiments and ablation studies to thoroughly verify the impact and effect of the proposed MetaCorrection framework. 
\end{itemize}

\vspace{-0.4cm}
\section{Related Work}
\vspace{-0.2cm}
\subsection{UDA in Semantic Segmentation}
\vspace{-0.2cm}
Unsupervised domain adaptation (UDA) aims to bridge the distribution gap between the labeled source domain and unlabeled target domain, thus improving the generalization capability of the learned models on the target data. The general idea of UDA semantic segmentation methods is to perform domain alignment through adversarial learning \cite{chen2019synergistic, chen2019learning, chen2019crdoco, choi2019self, du2019ssf, kim2020learning, li2019bidirectional, luo2019taking, tsai2018learning, tsai2019domain} or utilize self-training strategy on target samples to adapt the segmentation models  \cite{chen2019domain, iqbal2020mlsl, lian2019constructing, mei2020instance, pan2020unsupervised, sakaridis2019guided, zhang2019category, zheng2020rectifying, zhu2020improving, zou2019confidence, zou2018unsupervised}. We briefly review some typical works in the following parts.

\textbf{Adversarial Learning} based UDA semantic segmentation models usually contain two networks \cite{tsai2018learning}. One network behaves as a generator to obtain the segmentation maps for source and target inputs, while the other network serves as a discriminator to derive domain predictions. The generator intends to fool the discriminator to ensure the cross-domain alignment of feature \cite{chen2019synergistic, du2019ssf} or output \cite{kim2020learning, luo2019taking, tsai2018learning, tsai2019domain} levels. Other methods tried to narrow down the domain gap at the input level via image style transformation \cite{chen2019learning, chen2019crdoco, choi2019self, li2019bidirectional}. However, these domain alignment methods induced by adversarial learning ignored the domain-specific information and could not guarantee the discriminative ability for semantic segmentation \cite{wang2020unsupervised, zheng2020rectifying}.

\begin{figure*}
\centering
\includegraphics[width=142mm]{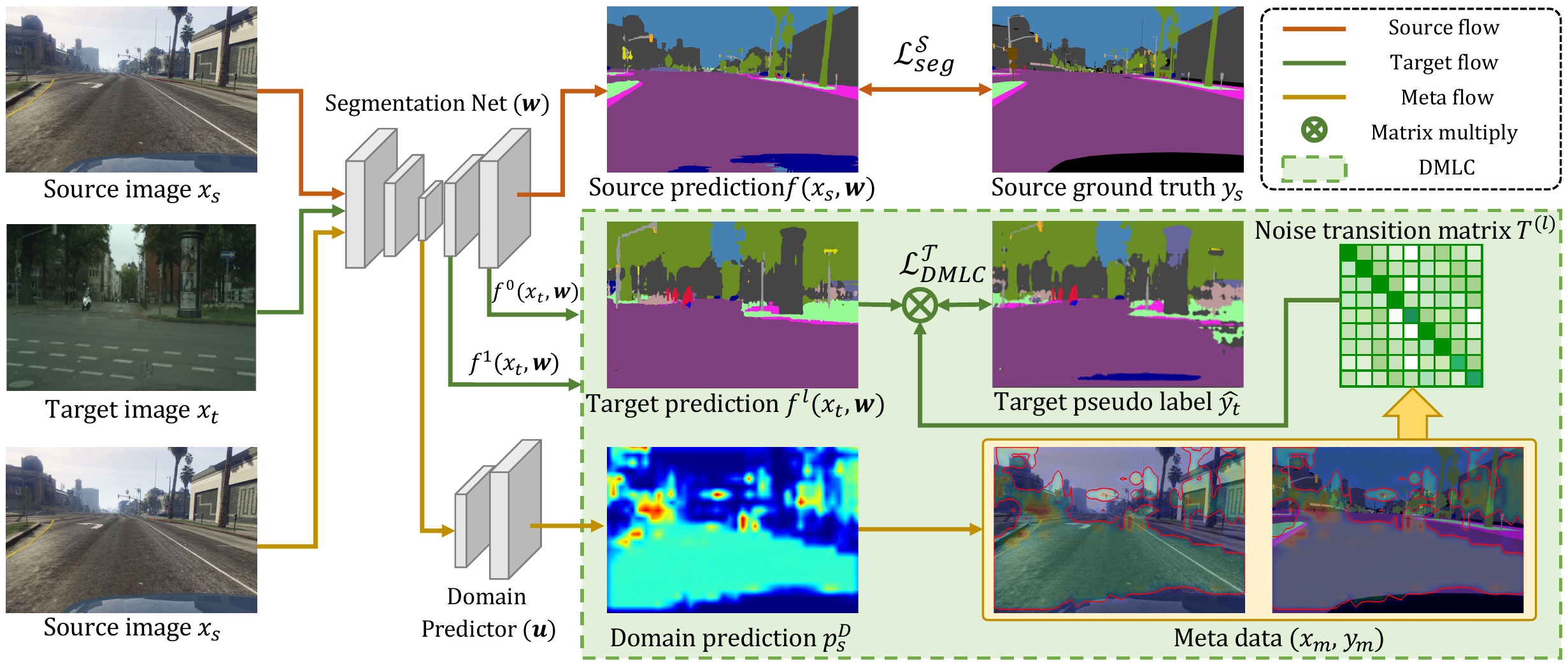}
\caption{The proposed MetaCorrection framework contains a segmentation net and a domain predictor. Both source and target images are passed through the segmentation net to perform semantic segmentation. The source data is supervised by the loss between prediction and the corresponding ground truth label, while the supervision signals of noisy pseudo-labeled target data are corrected by the learnable NTMs. Domain predictor is introduced to select domain-invariant source pixels for the guidance of NTM estimation.}
\label{fig:Framework}
\vspace{-0.3cm}
\end{figure*}

\textbf{Self-training.}
Another line of work for UDA semantic segmentation leverages the idea of self-training to adapt the segmentation model and learn the domain specific information \cite{mei2020instance, zhang2019category, zhu2020improving}. Previous methods \cite{chen2019domain, zou2019confidence} introduced entropy minimization to fully leverage the unlabeled target data for model training and encouraged the model to predict with high confidence score. Recently, increasing researchers investigated the problem of pseudo label noise in target domain by filtering out noisy samples with respect to confidence score or uncertainty \cite{iqbal2020mlsl, lian2019constructing, pan2020unsupervised, sakaridis2019guided, zheng2020rectifying, zou2018unsupervised}. Zou \textit{et. al.} \cite{zou2018unsupervised} proposed to threshold the argmax values of predictions and selected high-confidence pseudo-labeled samples. Zheng \textit{et. al.} \cite{zheng2020rectifying} utilized uncertainty estimation and enabled the dynamic threshold to obtain rectified pseudo labels. However, these methods only involved confident samples for training, which may result in biased prediction in minor classes and cannot distinguish confused categories. Moreover, when the noise ratio is high (at the early stage of training procedure), these models filtered out a large number of target samples, which may lose useful information in omitted samples. In this paper, we model the noise distribution of pseudo labels in target domain with a learnable matrix that encodes inter-class noise transition relationship, and propose a DMLC strategy to adaptively distill knowledge from all samples. 

\vspace{-0.1cm}
\subsection{Deep Learning with Noisy Labels}
\vspace{-0.1cm}
Many efforts have been devoted to tackling the noisy label problem in DNN training and can be roughly categorized into three typical strategies: label correction \cite{arazo2019unsupervised, zhang2020distilling}, sample reweighting \cite{jiang2018mentornet, li2017self, shu2019meta}, and loss correction \cite{goldberger2016training, hendrycks2018using, patrini2017making, shu2020meta, wang2020training}. 
Zhang \textit{et. al.} \cite{zhang2020distilling} introduced the meta-learning algorithm to conducted a dynamic linear combination of noisy label and prediction from DNN, thereby refurbishing noisy labels. Li \textit{et. al.} \cite{shu2020meta} progressively incorporated increasing samples in an easy-to-hard manner to enable a robust model trained with noisy samples. Authors in \cite{jiang2018mentornet, shu2020meta} utilized multiple layer perception network to automatically assign large weighting factor for easy samples. Goldberger \textit{et. al.} \cite{goldberger2016training} estimated the noise pattern through embedding a noise adaptation layer in DNN model. Wang \textit{et. al.} \cite{wang2020training} leveraged a small set of trusted clean-labeled samples to estimate the NTM for loss correction. Nonetheless, these strategies were designed for fully-supervised whole-image classification, and could not be directly incorporated into UDA  semantic segmentation. Our work represents the first effort to exploit the loss correction with an effective meta-learning strategy for self-training based UDA. 

\section{Methodology}
\vspace{-0.1cm}
\subsection{Preliminaries}
\vspace{-0.1cm}
We focus on the problem of UDA in semantic segmentation. In the source domain, we have access to source images $X_{\mathcal{S}} = \{x_{s} \in \mathbb{R}^{H\times W\times 3}\}_{s\in \mathcal{S}}$ and the corresponding pixel-wise one-hot labels $Y_{\mathcal{S}} =\{y_{s} \in \{0, 1\}^{H\times W\times C}\}_{s\in \mathcal{S}}$, while only target images $X_{\mathcal{T}} = \{x_{t} \in \mathbb{R}^{H\times W\times 3}\}_{t\in \mathcal{T}}$ are available in the target domain. Note that $H$, $W$, $C$ denote the height, width of images and the number of classes, respectively. The goal is to learn a segmentation net $f(\cdot)_\mathbf{w}$ that can correctly categorize pixels for target data $X_{\mathcal{T}}$. Self-training based methods  \cite{iqbal2020mlsl, lian2019constructing, pan2020unsupervised, sakaridis2019guided, zhang2019category, zhu2020improving, zou2019confidence} regarded pseudo labels of target data as learnable hidden variables, $\widehat{Y}_{\mathcal{T}} =\{\widehat{y}_{t}\}_{t\in \mathcal{T}}=\{\arg \max f({x}_{t})_\mathbf{w}\}_{t\in \mathcal{T}}$, and utilized them as approximate ground truth labels for model training. Then cross-entropy loss over the source and target dataset for self-training can be defined as
\begin{equation}
\begin{aligned}
\begin{split}
\mathcal{L}_{ST}&=\mathcal{L}_{seg}^{\mathcal{S}}(X_{\mathcal{S}}, Y_{\mathcal{S}}) + \mathcal{L}_{seg}^{\mathcal{T}}(X_{\mathcal{T}}, \widehat{Y}_{\mathcal{T}}) \\
&=-\sum_{s\in \mathcal{S}} {y}_{s} \log f\left({x}_{s}, \mathbf{w}\right)-\sum_{t\in \mathcal{T}} \widehat{y}_{t} \log f\left({x}_{t}, \mathbf{w}\right). 
\label{ST}  
\end{split}
\end{aligned}
\end{equation}
By minimizing the empirical risk of target data with respect to the estimated pseudo label $\widehat{y}_{t}$, the optimized model thus can be discriminatively adapted to the target domain.

\subsection{Self-training with Loss Correction} 
Jointly optimizing the model and estimating pseudo labels for target data is difficult as the accuracy of generated pseudo labels cannot be guaranteed. The noise in pseudo labels could deteriorate the performance of existing self-training based methods, and result in unstable training and biased predictions. A feasible way is to enhance the noise tolerance property of target domain risk minimization via loss correction. To incorporate loss correction, we assume that the generated pseudo labels $\widehat{Y}_{t}$ can be bridged to the ground truth labels  ${Y_{t}}$ via an underlying noise transition matrix (NTM) $T \in [0, 1]^{C\times C}$, which specifies the probability of ground truth label $j$ flipping to noisy label $k$ by $T_{jk} = p(\widehat{y}_{t} = k\mid {y}_{t} = j)$. 

If we directly optimize the segmentation net $f(\cdot)_{\mathbf{w}}$ on the noisy pseudo-labeled taget data, we would obtain a class posterior probability for noisy label $p(\widehat{y}_{t} = k \mid x_{t})$. NTM bridges the posterior for noisy label $p(\widehat{y}_{t} = k \mid x_{t})$ and the class probability for ground truth label via: 
\vspace{-0.2cm}
\begin{equation}
\begin{aligned}
\begin{split}
p(\widehat{y}_{t} = k \mid x_{t}, \mathbf{w})&=\sum_{j=1}^{C} T_{jk} p({y}_{t}=j \mid x_{t}, \mathbf{w}),\\
\Rightarrow p(\widehat{y}_{t} \mid x_{t}, &\mathbf{w})= p({y}_{t} \mid x_{t}, \mathbf{w})T.
\label{NTM}
\end{split}
\end{aligned}
\vspace{-0.1cm}
\end{equation}
Given the NTM ($T$), we modify and correct the self-training loss $\mathcal{L}_{seg}^{\mathcal{T}}(X_{\mathcal{T}})$ in Eq. (\ref{ST}) with respect to noisy pseudo-labeled target data as 
\begin{equation}
\vspace{-0.1cm}
\mathcal{L}_{LC}^{\mathcal{T}}(X_{\mathcal{T}}, \widehat{Y}_{\mathcal{T}})=-\sum_{t\in \mathcal{T}} \widehat{y}_{t} \log [f\left({x}_{t}, \mathbf{w}\right) T]. 
\label{LC}
\vspace{-0.1cm}
\end{equation}

This corrected loss function encourages the similarity between noise adapted posterior class probabilities and the noisy pseudo labels. It is obvious that once the NTM is obtained, we can recover the desired estimation of class posterior probability $p({y}_{t}|x_{t}, \mathbf{w})$ by the softmax output $f\left({x}_{t}, \mathbf{w}\right)$ even training the segmentation model with noisy data. 

\subsection{Domain-aware Meta Loss Correction (DMLC)} 
The effectiveness of loss correction methods highly depends on the estimation of NTM ($T$). Some previous attempts constructed $T$ with a strong assumption on the noise type \cite{hendrycks2018using, patrini2017making}, which impeded the generalization capability of model to complicated label noise. Other works required a set of clean labeled data to guide the estimation of $T$ \cite{hendrycks2018using, shu2020meta, wang2020training}. For example, Gold Loss Correction \cite{hendrycks2018using} utilizes the mean probability of all samples in the clean set categorized to class $i$ to approximate $T$. This requirement makes the loss correction algorithm infeasible to be applied to unsupervised learning task directly. 

To heuristically explore the inter-class noise transition probabilities, we devise a Domain-aware Meta-learning strategy to enable Loss Correction (DMLC) for UDA task. DMLC alternatively estimates $T$ by minimizing the empirical risk on the domain-invariant meta data with clean labels and optimizes the segmentation net with supervision signal corrected by previously approximated $T$ on the unlabeled target data. The key idea of DMLC lies in the estimation of $T$, and we first construct a set of meta data set $\{X_{\mathcal{M}}, M_{\mathcal{M}} \} = \{x_{m}, y_{m}\}_{m\in \mathcal{M}}$ with clean labels, representing the meta-knowledge of underlying label distribution of clean samples. Due to the lack of annotation in target domain, we attempt to select domain-invariant source data to construct such a meta data set. In particular, an additional pre-trained domain predictor $g(\cdot)_{\mathbf{u}}$ and a threshold coefficient $\tau$ are introduced to sample target-like source pixels as the meta data, as illustrated in Figure \ref{fig:Framework}. Only those samples with domain predictions $g(x_{s}, \mathbf{u})$ larger than $\tau$ are involved in meta-learning procedure. With the constructed meta data set, NTM can be updated to $T^{*}$ via:

\vspace{-0.4cm}
\begin{small}
\begin{equation}
\begin{aligned}
\begin{split}
T^{*}=\underset{T \in[0,1]^{c \times c}}{\arg \min } -\sum_{m\in \mathcal{M}} {y}_{m} \log f({x}_{m}, {\mathbf{w}(T)}^{*}), \\
where~\mathbf{w}(T)^{*}=\underset{\mathbf{w}}{\arg \min } -\sum_{t\in \mathcal{T}} \widehat{y}_{t} \log [f\left({x}_{t}, \mathbf{w}\right) T],
\label{DMLC1}
\end{split}
\end{aligned}
\vspace{-0.1cm}
\end{equation}
\end{small}where $\mathbf{w}(T)^{*}$ represents the optimal segmentation net with the minimal corrected loss on the noisy pseudo-labeled target data, and the updated $T^{*}$ should minimize the empirical loss on meta data with the optimal segmentation net \cite{shu2019meta, shu2020meta, wang2020training}. Intuitively, during the optimization procedure of segmentation net, it is difficult to distinguish hard samples and noisy samples since both of them can produce large loss values, {leading to the overfitting of noise}. Guiding the estimation of $T$ via risk minimization on meta data set, our method can avoid the wrong supervision signals and boost the generalization ability of adapted model. 

With the estimated $T^{*}$, the noisy pseudo-labeled target data can be effectively utilized to optimize the segmentation net. Jointly optimizing the segmentation net on source and target data, the proposed DMLC model can be extended from self-training in Eq. (\ref{ST}) and our objective function for UDA can be formulated as: 
\vspace{-0.1cm}
\begin{equation}
\begin{aligned}
\begin{split}
&\mathcal{L}_{DMLC} = \mathcal{L}_{seg}^{\mathcal{S}}(X_{\mathcal{S}}, Y_{\mathcal{S}}) +\mathcal{L}_{DMLC}^{\mathcal{T}} (X_{\mathcal{T}}, \widehat{Y}_{\mathcal{T}}) \\
&= -\sum_{s\in \mathcal{S}} {y}_{s} \log f\left({x}_{s}, \mathbf{w}\right) -\sum_{t\in \mathcal{T}} \widehat{y}_{t} \log [f\left({x}_{t}, \mathbf{w}\right) T^{*}]. 
\label{DMLC2}
\end{split}
\end{aligned}
\vspace{-0.1cm}
\end{equation}Jointly optimizing NTM associated with the segmentation net, the proposed DMLC can simultaneously estimate the noise distribution in pseudo labels and perform loss correction to target low segmentation error on target data. 

\subsection{Alternating Optimization for DMLC}
\begin{figure}[t]
\centering
\includegraphics[width=84mm]{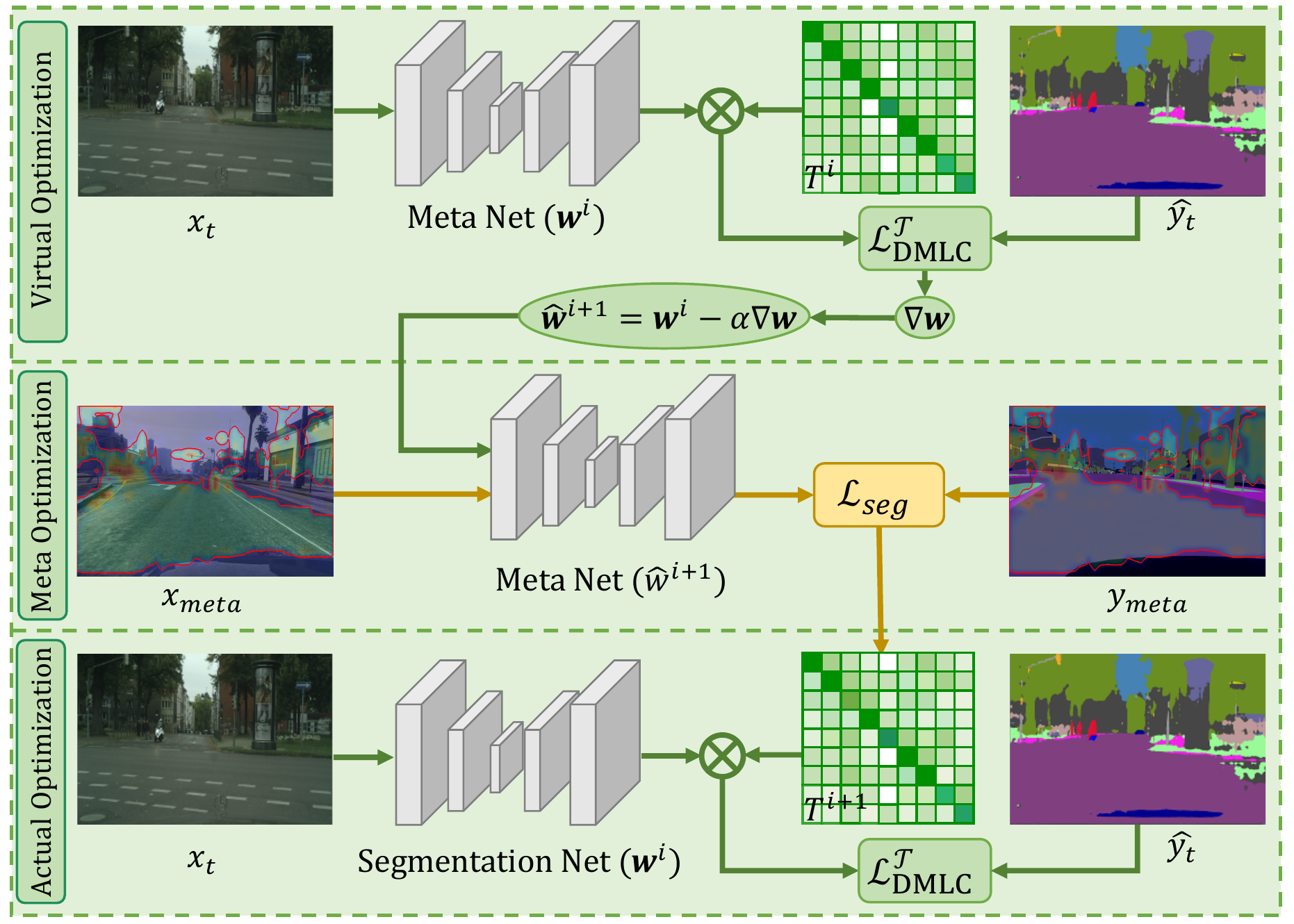}
\caption{Illustration of alternating optimization strategy for DMLC, including three steps: virtual optimization, meta optimization and actual optimization. }
\label{fig:MetaNT}
\vspace{-0.2cm}
\end{figure}

To synergically optimize the NTM ($T$) and the segmentation net $f(\cdot)\mathbf{w}$, we adopt an alternating optimization strategy, and the training procedure consists of three steps: \textit{virtual optimization, meta optimization, actual optimization}, as in Figure \ref{fig:MetaNT}. The virtual and meta optimizations aim to optimize $T$, and the actual optimization is to update parameters in the segmentation net with fixed $T$. 

During virtual optimization step (Figure \ref{fig:MetaNT} (a)), a meta net is copied from segmentation net with parameters $\mathbf{w}^{i}$, and a mini-batch of target images is forwarded  through the meta net. Then the parameters in the meta net are updated by moving the current $\mathbf{w}^{i}$ along the gradient descent direction of corrected loss function with learning rate $\gamma_{v}$: 
\vspace{-0.1cm}
\begin{equation}
\hat{\mathbf{w}}^{i+1}(T^{i})=\mathbf{w}^{i} + \gamma_{v} \nabla_{\mathbf{w}} \sum_{t\in \mathcal{T}} \widehat{y}_{t} \log [f\left({x}_{t}, \mathbf{w}^{i}\right) T^{i}]. 
\label{Op_DMLC1}
\vspace{-0.1cm}
\end{equation}
Note that this is a `virtual' step, indicating parameters in segmentation net are not actually updated to $\hat{\mathbf{w}}^{i+1}$.

Similar to well-known MAML \cite{finn2017model} with second-order back-propagation, in the meta optimization step (Figure \ref{fig:MetaNT} (b)), we update $T$ by minimizing the  cross entropy loss on meta data with the feedback from the updated parameters $\hat{\mathbf{w}}^{i+1}$ as follows: 
\vspace{-0.1cm}
\begin{equation}
\widetilde{T}^{i+1}=T^{i} + \gamma_{m} \nabla_{T} \sum_{m\in \mathcal{M}} {y}_{m} \log f({x}_{m}, \hat{\mathbf{w}}^{i+1}(T^{i})),
\label{Op_DMLC2}
\vspace{-0.1cm}
\end{equation}
where $\gamma_{m}$ is the learning rate of meta optimization. The intuition behind the meta optimization step is to obtain an optima of $\widetilde{T}^{i+1}$ with a low empirical risk and a high generalization ability. After the back propagation updating parameters, $\widetilde{T}^{i+1}$ may contains negative values. Therefore, we first utilize $\widetilde{T}^{i+1}=\max \left(\widetilde{T}^{i+1}, 0\right)$ to enable the non-negative matrix and then perform normalization along the row direction $T_{jk}^{i+1}={\widetilde{T}_{jk}^{i+1}}/{\sum \widetilde{T}_{j\cdot}^{i+1}}$ to ensure transition probabilities of class $j$ are summed to 1.

In the actual optimization step (Figure \ref{fig:MetaNT} (c)), the noisy pseudo-labeled target data and the labeled source data are simultaneously used to optimize the segmentation net via: 
\begin{equation}
\begin{aligned}
\begin{split}
{\mathbf{w}}^{i+1}=\mathbf{w}^{i} &+ \gamma_{a} \nabla_{\mathbf{w}}\sum_{s\in \mathcal{S}} {y}_{s} \log f\left({x}_{s}, \mathbf{w}\right)\\
&+\gamma_{a} \nabla_{\mathbf{w}}\sum_{t\in \mathcal{T}} \widehat{y}_{t} \log [f\left({x}_{t}, \mathbf{w}\right) T^{i+1}], 
\label{Op_DMLC4}
\end{split}
\end{aligned}
\vspace{-0.1cm}
\end{equation}
where $\gamma_{a}$ is the learning rate. Through the alternating optimization strategy, both the NTM ($T$) and the segmentation net $f(\cdot)_{\mathbf{w}}$ can be gradually ameliorated based on the optimal solution computed in the last step. 

\subsection{MetaCorrection}
Since low-level layers usually contain detailed features while high-level features often encode task-specific information, it may be beneficial to transfer the knowledge from the deep layer to guide the adaptation of low-level features. Previous deep supervision approaches \cite{chen2019domain, tsai2018learning} directly forces the low-level output to mimic the one-hot pseudo label computed from high-level output layer, which may bring about supervision bias and eliminate useful detailed information in low-level layer due to the capacity gap \cite{cho2019efficacy, liu2020metadistiller}. 

Aiming at solving the above problem, we incorporate the proposed DMLC to generate the matched and compatible supervision signal for the low-level features to enhance the adaptation. In particular, we introduce additional NTM for the loss correction and utilize the domain-aware meta-learning method to estimate the specific noise distribution for shallow supervisions, thereby bridging the gap between the low-level outputs and the pseudo label obtained from deep features. The overall training objective of our MetaCorrection framework can be extended from Eq. (\ref{DMLC2}) to be:
\begin{equation}
\begin{aligned}
\begin{split}
\mathcal{L}_{MC}= \mathcal{L}_{seg}^{\mathcal{S}}(X_{\mathcal{S}}, Y_{\mathcal{S}}) 
+ \sum_{l} \alpha_{l} \mathcal{L}_{DMLC}^{\mathcal{T} (l)}(X_{\mathcal{T}}, \widehat{Y}_{\mathcal{T}}), 
\label{ML_DMLC}
\end{split}
\end{aligned}
\vspace{-0.1cm}
\end{equation}
where $l$ indicates the level used for DMLC and $\alpha_{l}$ is the weighting factor for supervision signal of $l^{th}$ layer. Note that $l=0$ denotes the output layer. 

With the proposed MetaCorrection framework, NTMs embedded in the segmentation net are capable of assessing the individual noise distributions in pseudo labels for different layers. Moreover, our MetaCorrection framework can obtain corrected loss functions in a data-driven manner, and these generated compatible supervision signals for different levels of features can further boost the learning of model.

\begin{table*}[tp!]
\centering
\caption{Results of adapting GTA5 to CityScapes. The mechanism `AL' and `ST' stand for adversarial learning and self-training. }
\vspace{+0.1cm}
\label{GTA5Cityscapes}
\scalebox{0.67}{\begin{tabular}{c c | c  c  c  c  c  c  c  c  c  c  c  c  c  c  c  c  c  c  c | c}
\ChangeRT{1.5pt}
&\multicolumn{21}{c}{GTA5 $\rightarrow$ CityScapes}\\

\ChangeRT{1.0pt}
Methods&\rotatebox{90}{mech.}&\rotatebox{90}{road}&\rotatebox{90}{sidewalk}&\rotatebox{90}{building}&\rotatebox{90}{wall}&\rotatebox{90}{fence}&\rotatebox{90}{pole}&\rotatebox{90}{traffic lgt}&\rotatebox{90}{traffic sgn}&\rotatebox{90}{veg.}&\rotatebox{90}{terrain}&\rotatebox{90}{sky}&\rotatebox{90}{person}&\rotatebox{90}{rider}&\rotatebox{90}{car}&\rotatebox{90}{truck}&\rotatebox{90}{bus}&\rotatebox{90}{train}&\rotatebox{90}{motor}&\rotatebox{90}{bike}&{mIoU}\\

\hline
AdaptSegNet \cite{tsai2018learning}&AL&86.5&36.0&79.9&23.4&23.3&23.9&35.2&14.8&83.4&33.3&75.6&58.5&27.6&73.7&32.5&35.4&3.9&30.1&28.1&42.4 \\

PatchAlign \cite{tsai2019domain}&AL&92.3&51.9&82.1&29.2&25.1&24.5&33.8&33.0&82.4&32.8&82.2&58.6&27.2&84.3&33.4&46.3&2.2&29.5&32.3&46.5 \\

LTIR \cite{kim2020learning}&AL&\textbf{92.9}&55.0&85.3&34.2&31.1&34.9&40.7&34.0&85.2&\underline{40.1}&{87.1}&61.0&31.1&82.5&32.3&42.9&0.3&36.4&46.1&50.2 \\

\hline
CBST \cite{zou2018unsupervised}&ST&91.8&53.5&80.5&32.7&21.0&34.0&28.9&20.4&83.9&34.2&80.9&53.1&24.0&82.7&30.3&35.9&16.0&25.9&42.8&45.9 \\

CRST \cite{zou2019confidence}&ST&91.0&55.4&80.0&33.7&21.4&37.3&32.9&24.5&85.0&34.1&80.8&57.7&24.6&84.1&27.8&30.1&26.9&26.0&42.3&47.1 \\

MaxSquare \cite{chen2019domain}&ST&89.4&43.0&82.1&30.5&21.3&30.3&34.7&24.0&85.3&39.4&78.2&\textbf{63.0}&22.9&84.6&\underline{36.4}&43.0&5.5&34.7&33.5&46.4 \\

MLSL \cite{iqbal2020mlsl}&ST&89.0&45.2&78.2&22.9&27.3&\underline{37.4}&\textbf{46.1}&\underline{43.8}&82.9&18.6&61.2&60.4&26.7&85.4&35.9&44.9&\underline{36.4}&\textbf{37.2}&\textbf{49.3}&49.0 \\

PyCDA \cite{lian2019constructing}&ST&90.5&36.3&84.4&32.4&28.7&34.6&36.4&31.5&\textbf{86.8}&37.9&78.5&62.3&21.5&\underline{85.6}&27.9&34.8&18.0&22.9&\textbf{49.3}&47.4 \\

IntraDA \cite{pan2020unsupervised}&ST&90.6&37.1&82.6&30.1&19.1&29.5&32.4&20.6&\underline{85.7}&\textbf{40.5}&79.7&58.7&31.1&\textbf{86.3}&31.5&\underline{48.3}&0.0&30.2&35.8&46.3 \\

CAG-UDA \cite{zhang2019category}&ST&90.4&51.6&83.8&34.2&27.8&\textbf{38.4}&25.3&\textbf{48.4}&85.4&38.2&78.1&58.6&\textbf{34.6}&84.7&21.9&42.7&\textbf{41.1}&29.3&37.2&50.2 \\


\hline
Source only&--&75.8&16.8&77.2&12.5&21.0&25.5&30.1&20.1&81.3&24.6&70.3&53.8&26.4&49.9&17.2&25.9&6.5&25.3&36.0&36.6\\

Ours (single DMLC)&ST&92.5&\underline{55.1}&\underline{85.9}&\underline{36.9}&\underline{32.4}&34.7&41.4&37.0&85.3&37.8&\underline{87.4}&{62.7}&\underline{31.8}&{84.5}&\textbf{36.8}&48.2&2.2&34.3&47.3&\underline{51.2} \\

Ours (MetaCorrection)&ST&\underline{92.8}&\textbf{58.1}&\textbf{86.2}&\textbf{39.7}&\textbf{33.1}&36.3&\underline{42.0}&38.6&85.5&37.8&\textbf{87.6}&\underline{62.8}&31.7&84.8&35.7&\textbf{50.3}&2.0&\underline{36.8}&\underline{48.0}&\textbf{52.1} \\
\ChangeRT{1.5pt}
\end{tabular}}
\vspace{-0.4cm}
\end{table*}

\section{Experiments}
\subsection{Datasets}
We evaluate the performance of our methods on two challenging synthetic-to-real UDA semantic segmentation tasks and a medical image segmentation task. Two synthetic datasets, GTA5 \cite{richter2016playing} and SYNTHIA \cite{ros2016synthia}, and a real dataset, CityScapes \cite{cordts2016cityscapes}, are utilized to perform UDA synthetic-to-real semantic segmentation tasks, including two scenarios: GTA5$\rightarrow$CityScapes and SYNTHIA$\rightarrow$CityScapes. Moreover, two public prostate MRI datasets are adopted to perform UDA from Decathlon \cite{simpson2019large} to NCI-ISBI13 \cite{NCIISBI}. 

\textbf{GTA5} contains 24,966 images captured from a video game. Pixel-wise annotations with 33 classes are provided, but only 19 classes are utilized for compatibility with CityScapes. 
\textbf{SYNTHIA} consists of 9,400 synthetic images, and annotations with 16 classes are used for adaptation.
\textbf{CityScapes}  is a real-world semantic segmentation dataset collected in driving scenarios. Training set, including 2,975 unlabeled images, is regarded as the target domain data for training. Evaluations are performed on 500 validation images with manual annotations. 
\textbf{Decathlon} is a comprehensive medical image segmentation dataset, including 32 prostate MRI scans obtained from 3T (Siemens TIM). Annotations outline the peripheral zone (PZ) and transition zone (TZ). \textbf{NCI-ISBI13} consists of 40 labeled prostate MRI scans obtained from 1.5 T (Philips Achieva). We utilized 30 training scans as unlabeled target data to perform UDA training and other 10 scans for evaluation.

\subsection{Network and Training Details}
\vspace{-0.1cm}
\textbf{Segmentation Net} We adopt DeepLab-v2 \cite{chen2017deeplab} backbone with pre-trained ResNet-101 \cite{he2016deep} encoder as our segmentation net. Subsequently, Atrous Spatial Pyramid Pooling (ASPP) is employed after the last layer of encoder with dilated rates \{6, 12, 18, 24\}. Finally, an up-sampling layer along with a softmax operator is applied to obtain the final segmentation result with the matched size of input image. 

We construct the above-mentioned segmentation net and apply the NTM to the output layer as our single DMLC model. For the MetaCorrection framework, we additionally extract low-level feature maps from the $conv4$ layer of ResNet-101 and introduce an ASPP module as the auxiliary classifier with output $f^{1}\left({x}_{t}, \mathbf{w}\right)$. An additional NTM $T^{(1)}$ is incorporated to generate a compatible supervision signal for low-level output. 
$\alpha_{0}$ and $\alpha_{1}$ in Eq. (\ref{ML_DMLC}) are set as 1, 0.1. 

\textbf{Domain Predictor} The feature maps extracted from the encoder in the segmentation net are utilized for the pixel-level domain prediction. We adopt a similar structure with DCGAN \cite{radford2015unsupervised}, which is composed of five cascaded 4$\times$4 convolution layers with output channel numbers \{64, 128, 256, 512, 1\}. Then the domain prediction is obtained with the same resolution of the input image, and the threshold coefficient $\tau$ is set as 0.5 to construct the meta data set. 

\textbf{Implementation Details} Our methods are implemented with the PyTorch library on Nvidia Tesla V100. The Stochastic Gradient Descent is utilized as our optimizer, where the momentum is 0.9 and the weight decay is 1e-3. We adopt polynomial learning rate scheduling to optimize the segmentation net with the initial learning rate of $\gamma_{a}=2.5e-4$, the power of 0.9 and the maximum iteration number of 150000. For the updating of meta net, $\gamma_{v}=1e-4$ and $\gamma_{m}=0.11$ are set in our implementation. 

The performances of our methods in synthetic-to-real scenarios are evaluated by the widely utilized performance metrics, intersection-over-union (IoU) of each class and the mean IoU (mIoU). For the prostate zonal segmentation, $Dice$ scores for PZ, TZ and the whole prostate (WP) are employed to measure the accuracy of segmentation results. 

\begin{table*}[tp!]
\centering
\caption{Results of adapting SYNTHIA to CityScapes. mIoU* denotes the mean IoU of 13 classes, excluding the classes with $*$.}
\vspace{+0.1cm}
\label{SYNTHIACityscapes}
\scalebox{0.67}{\begin{tabular}{c  c | c  c  c  c  c  c  c  c  c  c  c  c  c  c  c  c |  c  c}
\ChangeRT{1.5pt}
&\multicolumn{19}{c}{SYNTHIA $\rightarrow$ CityScapes}\\

\ChangeRT{1.0pt}
Methods&\rotatebox{90}{mech.}&\rotatebox{90}{road}&\rotatebox{90}{sidewalk}&\rotatebox{90}{building}&\rotatebox{90}{wall*}&\rotatebox{90}{fence*}&\rotatebox{90}{pole*}&\rotatebox{90}{light}&\rotatebox{90}{sign}&\rotatebox{90}{veg.}&\rotatebox{90}{sky}&\rotatebox{90}{person}&\rotatebox{90}{rider}&\rotatebox{90}{car}&\rotatebox{90}{bus}&\rotatebox{90}{mbike}&\rotatebox{90}{bike}&{mIoU}&{mIoU*}\\

\hline
AdaptSegNet \cite{tsai2018learning}&AL&84.3&42.7&77.5&--&--&--&4.7&7.0&77.9&82.5&54.3&21.0&72.3&32.2&18.9&32.3&--&46.7 \\

PatchAlign \cite{tsai2019domain}&AL&82.4&38.0&78.6&8.7&0.6&26.0&3.9&11.1&75.5&84.6&53.5&21.6&71.4&32.6&19.3&31.7&40.0&46.5 \\

LTIR \cite{kim2020learning}&AL&\textbf{92.6}&\textbf{53.2}&79.2&--&--&--&1.6&7.5&78.6&84.4&52.6&20.0&82.1&34.8&14.6&39.4&--&49.3\\

\hline
CBST \cite{zou2018unsupervised}&ST&68.0&29.9&76.3&10.8&1.4&33.9&22.8&29.5&77.6&78.3&60.6&\underline{28.3}&81.6&23.5&18.8&39.8&42.6&48.9 \\

CRST \cite{zou2019confidence}&ST&67.7&32.2&73.9&10.7&1.6&\textbf{37.4}&22.2&\underline{31.2}&80.8&80.5&60.8&\textbf{29.1}&82.8&25.0&19.4&45.3&43.8&50.1 \\

MaxSquare \cite{chen2019domain}&ST&82.9&40.7&\underline{80.3}&10.2&0.8&25.8&12.8&18.2&82.5&82.2&53.1&18.0&79.0&31.4&10.4&35.6&41.4&48.2 \\

MLSL \cite{iqbal2020mlsl}&ST&59.2&30.2&68.5&\textbf{22.9}&1.0&\underline{36.2}&\textbf{32.7}&{28.3}&\textbf{86.2}&75.4&\textbf{68.6}&27.7&82.7&26.3&\textbf{24.3}&\textbf{52.7}&\underline{45.2}&51.0 \\

PyCDA \cite{lian2019constructing}&ST&75.5&30.9&83.3&\underline{20.8}&0.7&32.7&\underline{27.3}&\textbf{33.5}&\underline{84.7}&85.0&64.1&25.4&\underline{85.0}&45.2&21.2&32.0&\textbf{46.7}&\textbf{53.3} \\

IntraDA \cite{pan2020unsupervised}&ST&84.3&37.7&79.5&5.3&0.4&24.9&9.2&8.4&80.0&84.1&57.2&23.0&78.0&\textbf{38.1}&20.3&36.5&41.7&48.9 \\

CAG-UDA \cite{zhang2019category}&ST&84.7&40.8&81.7&7.8&0.0&{35.1}&13.3&22.7&84.5&77.6&\underline{64.2}&{27.8}&80.9&19.7&\underline{22.7}&\underline{48.3}&44.5&51.5 \\
 

\hline
Source only&--&55.6&23.8&74.6&9.2&0.2&24.4&6.1&12.1&74.8&79.0&55.3&19.1&39.6&23.3&13.7&25.0&33.5&38.6\\

Ours (single DMLC)&ST&\underline{92.3}&\underline{53.0}&80.2&7.7&\textbf{2.8}&26.9&11.4&8.1&83.1&\textbf{85.2}&58.9&20.5&\textbf{85.5}&35.9&21.0&41.8&44.6&52.1 \\

Ours (multi DMLC)&ST&\textbf{92.6}&52.7&\textbf{81.3}&8.9&\underline{2.4}&28.1&13.0&7.3&83.5&\underline{85.0}&60.1&19.7&84.8&\underline{37.2}&21.5&43.9&45.1&\underline{52.5} \\
\ChangeRT{1.5pt}
\end{tabular}}
\vspace{-0.3cm}
\end{table*}

\begin{table}[tp!] 
\caption{Results of adapting Decathlon to NCI-ISBI13.}
\vspace{+0.1cm}
\centering
\noindent
\scalebox{0.68}{\begin{tabular}{c c |  c  c  c}
\ChangeRT{1.5pt}
\xrowht{11pt}
Method&mech.&PZ ($Dice$)&TZ ($Dice$)&WP ($Dice$)\\
\ChangeRT{1.0pt}
CBST \cite{zou2018unsupervised}&ST&38.22&70.14&64.31\\
MRENT \cite{zou2019confidence}&ST&40.82&72.39&67.68\\
MaxSquare \cite{chen2019domain}&ST&37.45&69.61&63.34\\
\hline
Source only&--&28.48&52.57&47.56\\
Ours (single DMLC)&ST&\underline{42.03}&\underline{74.09}&\underline{69.38}\\
Ours (MetaCorrection)&ST&\textbf{43.25}&\textbf{74.31}&\textbf{70.87}\\
\ChangeRT{1.5pt}
\end{tabular}}
\label{Prostate}
\vspace{-0.4cm}
\end{table}

\subsection{Results on GTA5$\rightarrow$CityScapes}
\vspace{-0.1cm}
We first verify the effectiveness of our approachs in the GTA5$\rightarrow$Cityscapes scenario, and the corresponding comparison results are listed in Table \ref{GTA5Cityscapes} with the first and second best results highlighted in bold and underline. For a fair comparison, all the competed models adopt DeepLab-v2 backbone network with pre-trained ResNet-101 as encoder. Overall, our MetaCorrection framework surpasses all other models with a promising mIoU of 52.1\%, outperforming the model trained only on the source data by a significant increment of 15.5\% in mIoU. Compared with domain alignment methods \cite{tsai2018learning, tsai2019domain, kim2020learning}, the proposed method shows superior performance. For example, PatchAlign \cite{tsai2019domain} leverages the patch-level information to encourage the domain alignment, yielding 46.5\% mIoU, which is inferior to our approach. Our MetaCorrection model, as a self-training based method, also outperforms other related pseudo label learning works \cite{zou2018unsupervised, chen2019domain, iqbal2020mlsl, lian2019constructing, pan2020unsupervised, zhang2019category}, demonstrating the effectiveness of the proposed method in alleviating the noise problem. Moreover, the proposed methods also show superior performance in terms of the per class IoU score, especially in the minor categories (e.g., `motor') and ambiguous (e.g., `road' and `sidewalk') categories. 

\subsection{Results on SYNTHIA $\rightarrow$ CityScapes}
\vspace{-0.1cm}
We then utilize SYNTHIA as the source domain data and display comparison results of our methods and other state-of-the-art methods \cite{tsai2018learning, tsai2019domain, kim2020learning, zou2018unsupervised, chen2019domain, iqbal2020mlsl, lian2019constructing, pan2020unsupervised, zhang2019category} on the validation set of Cityscapes, as listed in the Table \ref{SYNTHIACityscapes}. We consider the IoU and mIoU of both the 16 classes and a subset of 13 classes following the standard experimental setting \cite{pan2020unsupervised}. Since the domain shift is more evident in this scenario, the performance is slightly worse. Our MetaCorrection framework still achieves promising results in comparison to other competed methods. Specifically, the proposed method achieves 45.1\% mIoU of 16 categories and 52.5\% mIoU$^{*} $of 13 categories. 

\subsection{Results on Decathlon $\rightarrow$ NCI-ISBI13}
Domain discrepancy is common in clinical practice, e.g., MRIs obtained from different scanners and sites. Hence, we further assess the performance of our methods on UDA prostate segmentation, and the quantitative comparison results are listed in Table \ref{Prostate}. It is observed that our methods exhibit superior segmentation performance in comparison to the self-training based methods, CBST \cite{zou2018unsupervised}, MRENT \cite{zou2019confidence}, MaxSquare \cite{chen2019domain}. For example, both single DMLC and MetaCorrection methods outperform CBST with increments of 5.07\%, 6.56\% in WP $Dice$ score. This observation proves the impact of our approaches in medical image analysis. 

\subsection{Ablation Study}
\textbf{Ablation Experiments.}
To investigate the effects of individual components of our proposed model, we design ablation studies under three adaptation scenarios (Table \ref{GTA5Cityscapes}, \ref{SYNTHIACityscapes}, \ref{Prostate}) and with three baseline models (Table \ref{DiffNoise}).  Compared with the `Source only' lower bound, our baseline network with single DMLC boosts the mIoU to 51.2\% with an increment of 14.6\% in GTA5 $\rightarrow$ CityScapes case. Then we introduce auxiliary supervision signals for low-level layers, which also contributes to the performance gain and increases the mIoU to 52.1\%, as in Table \ref{GTA5Cityscapes}. Moreover, incorporating DMLC to different baseline models, the MetaCorrection framework consistently improves the performance over single DMLC in terms of mIoU score, as in Table \ref{DiffNoise}. 

\begin{table}[tp!] 
\caption{Impact of different pseudo labels. `Pseudo Label' denotes we employ pseudo labels generated by the corresponding model.}
\vspace{+0.1cm}
\centering
\noindent
\scalebox{0.67}{\begin{tabular}{c | c | c  c }
\ChangeRT{1.5pt}
\xrowht{11pt}
Method&Pseudo Label&\begin{tabular}[c]{@{}c@{}}GTA5 $\rightarrow$ \\CityScapes\end{tabular}&$\Delta$\\
\ChangeRT{1.0pt}
AdaptSegNet \cite{tsai2018learning}&---&42.4&--\\
Self-training (MRENT \cite{zou2019confidence})&AdaptSegNet&45.1&2.7\\
Self-training (Threshold \cite{zou2018unsupervised})&AdaptSegNet&44.4&2.0\\
Self-training (Ucertainty \cite{zheng2020rectifying})&AdaptSegNet&46.1&3.7\\
Ours (single DMLC)&AdaptSegNet&45.9&3.5\\
Ours (MetaCorrection)&AdaptSegNet&\textbf{47.3}&\textbf{4.9}\\

\ChangeRT{1.0pt}
LTIR \cite{kim2020learning}&---&50.2&--\\
Self-training (MRENT \cite{zou2019confidence})&LTIR&50.6&0.4\\
Ours (single DMLC)&LTIR&51.2&1.0\\
Ours (MetaCorrection)&LTIR&\textbf{52.1}&\textbf{1.9}\\

\ChangeRT{1.0pt}
Source only&---&36.6&--\\
Self-training (MRENT \cite{zou2019confidence})&Source&39.6&3.0\\
Ours (single DMLC)&Source&43.8&7.2\\
Ours (MetaCorrection)&Source&\textbf{44.5}&\textbf{7.9}\\
\ChangeRT{1.5pt}
\end{tabular}}
\vspace{-0.1cm}
\label{DiffNoise}
\vspace{-0.3cm}
\end{table}

\begin{figure*}
\centering
\includegraphics[width=167mm]{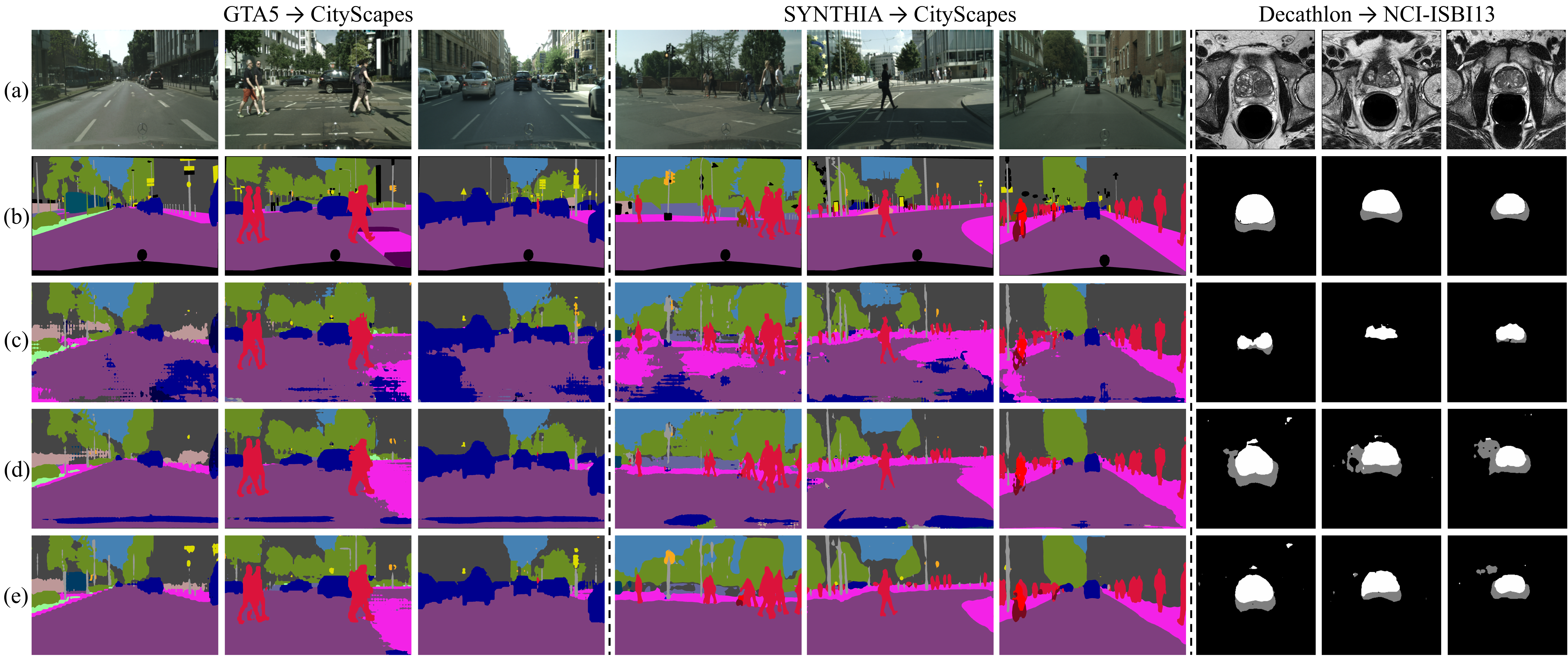}
\caption{Qualitative results of UDA semantic segmentation. (a) Target image,  (b) Ground truth, Predictions from (c) source only model, (d) self-training based MRENT model \cite{zou2019confidence}, (e) ours (MetaCorrection).}
\label{fig:seg}
\vspace{-0.3cm}
\end{figure*}

\textbf{Comparison with Self-training based UDA Models.}
To validate the effectiveness of DMLC, we compare it with three typical self-training based UDA models \cite{zheng2020rectifying, zou2019confidence, zou2018unsupervised}. As listed in Table \ref{DiffNoise}, the proposed MetaCorrection framework ($row$ 7) is superior to other self-training methods, including entropy minimization \cite{zou2019confidence} ($row$ 3), handcrafted threshold  \cite{zou2018unsupervised} ($row$ 4), uncertainty based rectification \cite{mei2020instance} ($row$ 5), yielding increments of 2.2\%, 2.9\%, 1.2\% mIoU. Through entropy minimization, considerable noisy labels inevitably result in unsatisfactory performance. Other self-training based methods \cite{zou2019confidence, zou2018unsupervised} filtered out noisy samples with respect to the confidence score and uncertainty, but lost useful information in those omitted samples. Therefore, the increments are owing to that our method can preserve all data distribution and distill effective information from all samples with learned NTM for unbiased self-training. 

\textbf{Robustness to Various Types of Noise.}
We further explore the robustness of our MetaCorrection framework under different types of noise. Specifically, we adopt AdaptSegNet \cite{tsai2018learning}, LTIR \cite{kim2020learning}, source only model to generate noisy pseudo labels and apply our MetaCorrection to mitigate the noise problem. As in Table \ref{DiffNoise}, these three models have significantly improved performance with the proposed MetaCorrection framework. For example, MetaCorrection ($row$ 7) improves the performance of AdaptSegNet ($row$ 2) from 42.4\% to 47.3\% mIoU. 

\begin{figure}
\centering
\includegraphics[width=83mm]{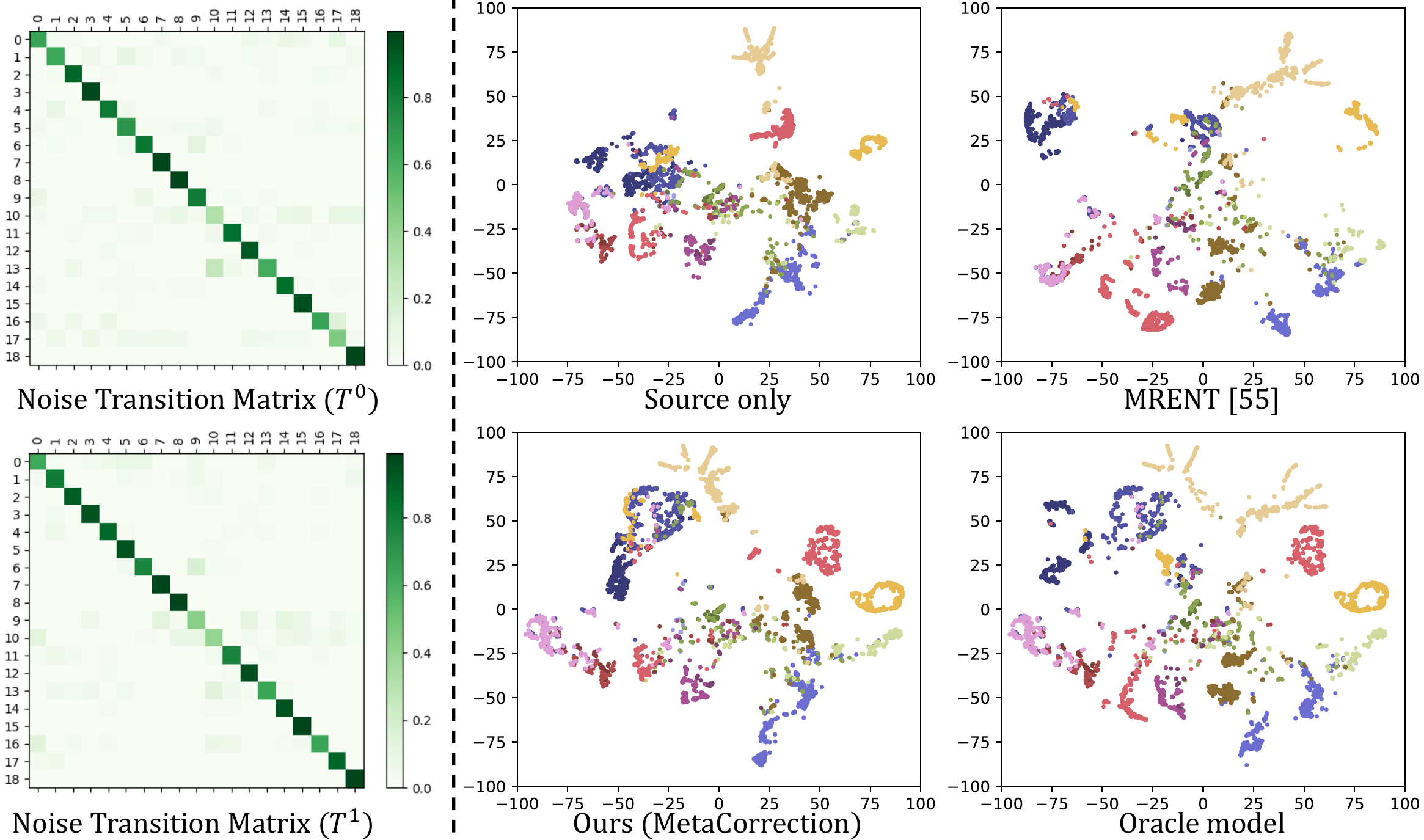}
\caption{Left: Visualization of NMTs $T^{(0)}$ and $T^{(1)}$. Right: The t-SNE visualization of embedded features.}
\label{fig:tsne}
\vspace{-0.4cm}
\end{figure}

\subsection{Visualization Results}
\vspace{-0.1cm}
\textbf{Segmentation Visualization.} 
As illustrated in Figure \ref{fig:seg}, we provide some typical qualitative segmentation results of target data on all three benchmarks. Obviously, the self-training method \cite{zou2019confidence} could significantly promote the performance in comparison to the source model. Besides, in contrast to the baseline self-training method with conventional entropy minimization, the proposed MetaCorrection framework has better scalability to confused categories (e.g., `rider' and `bike') and small-scale objectives (e.g., `traffic sgn'). We speculate the reason is that pseudo labels usually contain considerable noises in the minor categories and ambiguous categories. The proposed method rectifies the supervision signals and prevent such mistakes, leading to more reasonable segmentation predictions.

\textbf{NTM Visualization.} We visualize the learned NTMs of output layer ($T^{0}$) and shallow layer ($T^{1}$), as in Figure \ref{fig:tsne}. It is obvious that different layers exhibit variant noise transition probability, indicating the varying noise distributions of deep and shallow layers. The proposed MetaCorrection framework can generate matched supervision signals for individual layers to enhance the deep adaptation.

\textbf{Feature Visualization.} 
We use t-SNE \cite{maaten2008visualizing} to visualize the feature representations of source only model, self-training \cite{zou2019confidence}, our MetaCorrection and oracle model (i.e., fine-tuning the segmentation net with the labeled target data), as illustrated in Figure \ref{fig:tsne}. It is observed that our MetaCorrection model obtains the most matched feature distribution with that of the oracle model in comparison to source only and self-training \cite{zou2019confidence} models. This observation demonstrates that our method can provide correct supervision signal for target data through the learnable NTM. Moreover, our feature representations exhibit the clearest clusters compared with other baseline methods, revealing the discriminative capability of our method.

\vspace{-0.1cm}
\section{Conclusion}
\vspace{-0.1cm}
In this paper, we have proposed a MetaCorrection framework, where the Domain-aware Meta Loss Correction (DMLC) strategy is advanced for UDA in the context of semantic segmentation, aiming for addressing the noise problem in self-training based UDA methods. The DMLC incorporates a learnable noise transition matrix (NTM) to bridge the noisy pseudo labels and ground truth labels for loss correction of the target domain, and NTM is derived through the proposed domain-aware meta-learning strategy in a data-driven manner. The model-agnostic DMLC can be flexibly applied to other models and datasets. Moreover, we consider the capacity gap between deep and shallow layers, and provide compatible supervisions for different levels to ensure the deep adaptation of the proposed MetaCorrection. The experimental results demonstrate that our methods achieve superior results to state-of-the-art methods.

{\small
\bibliographystyle{ieee_fullname}
\bibliography{egbib}
}

\end{document}